\documentclass[10pt,twocolumn,letterpaper]{article}

\usepackage[pagenumbers]{cvpr} 

\usepackage{graphicx}
\usepackage{amsmath}
\usepackage{amssymb}
\usepackage{booktabs}
\usepackage{multicol}
\usepackage{multirow}
\usepackage{float}
\usepackage{color}

\usepackage[pagebackref,breaklinks,colorlinks]{hyperref}

\usepackage[capitalize]{cleveref}
\crefname{section}{Sec.}{Secs.}
\Crefname{section}{Section}{Sections}
\Crefname{table}{Table}{Tables}
\crefname{table}{Tab.}{Tabs.}

\def\cvprPaperID{10876} 
\def\confName{CVPR}
\def\confYear{2023}

\begin{document}

\title{TCNL: Transparent and Controllable Network Learning Via Embedding Human-Guided Concepts}

\author{Zhihao Wang\\
Beijing University of Posts and Telecommunications\\
Beijing, China \\
{\tt\small SH18540106@163.com}
\and
Chuang Zhu\\
Beijing University of Posts and Telecommunications\\
Beijing, China \\
{\tt\small czhu@bupt.edu.cn}
}
\maketitle

\begin{abstract}
Explaining deep learning models is of vital importance for understanding artificial intelligence systems, improving safety, and evaluating fairness. To better understand and control the CNN model, many methods for transparency-interpretability have been proposed. However, most of these works are less intuitive for human understanding and have insufficient human control over the CNN model. We propose a novel method, Transparent and Controllable Network Learning (TCNL), to overcome such challenges. Towards the goal of improving transparency-interpretability, in TCNL, we define some concepts for specific classification tasks through scientific human-intuition study and incorporate concept information into the CNN model. In TCNL, the shallow feature extractor gets preliminary features first. Then several concept feature extractors are built right after the shallow feature extractor to learn high-dimensional concept representations. The concept feature extractor is encouraged to encode information related to the predefined concepts. We also build the concept mapper to visualize features extracted by the concept extractor in a human-intuitive way. TCNL provides a generalizable approach to transparency-interpretability. Researchers can define concepts corresponding to certain classification tasks and encourage the model to encode specific concept information, which improves transparency-interpretability and the controllability of the CNN model. The datasets (including concept instance sets) for our experiments will be released for scientific research (\url{https://github.com/bupt-ai-cz/TCNL}).
\end{abstract}

\section{Introduction}
\label{sec:intro}

\begin{figure}[ht]
    \centering
    \includegraphics[width=0.9\linewidth]{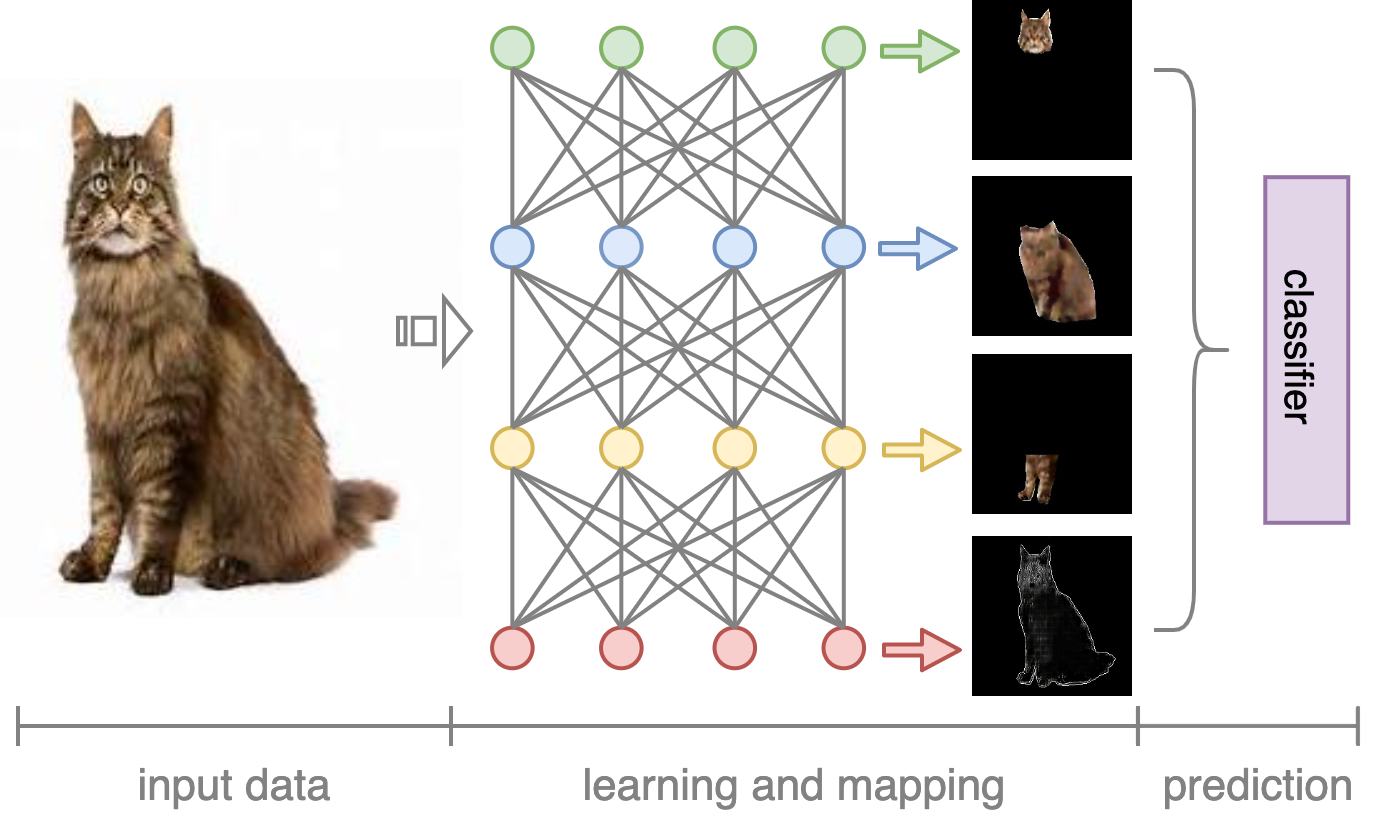}
    \caption{This figure takes the mammal classification task as an example to show the workflow of TCNL. In TCNL, the CNN model first extracts concept features, then maps them to concept instance images for better human understanding. Based on concept features, the model makes the final prediction. The learning process and inference process are transparent and controllable.}
    \label{figure:flow}
\end{figure}

Recently, the convolutional neural networks (CNN)~\cite{hinton, alexnet, resnet, inception}, have achieved excellent performance in various computer vision tasks such as image classification, object detection, and semantic segmentation. Besides the superior performance, the interpretability of the model plays a critical role in safety, fairness, and scientific research. Towards the goal of building the trusty artificial intelligence system, more and more scholars devote themselves to the study of the interpretability of the CNN.

Nowadays, there are two main types of interpretable algorithms. One is designed to improve the transparency of the CNN by adjusting the structure of the model, named as the transparency-interpretability method~\cite{mythos}. The other aims at giving a reasonable explanation for the decision of the CNN, named as the post-hoc interpretability method~\cite{mythos}.

Although some progress~\cite{XAIsurvey} has been made in the area of the interpretability of the CNN, some issues remain unsolved. Many transparency-interpretability works focus on improving the interpretability of the CNN by changing the structure of the model~\cite{entropy, part_whole, CSGCNN, shapleynet}. However, most of these works interpret the CNN model in a way that is less intuitive for human understanding~\cite{part_whole, shapleynet}. For post-hoc interpretability methods that give visual explanations of the CNN, many methods~\cite{cam, gradcam, scorecam, relevancecam} do the visualization by operating feature maps from a specific convolution layer. These methods try to explain decisions of the model in a linear way. Post-hoc interpretability methods work on an already trained model, therefore they can not change the fact that the CNN still lacks interpretability.

\begin{figure*}[ht]
    \centering
    \includegraphics[width=0.9\linewidth]{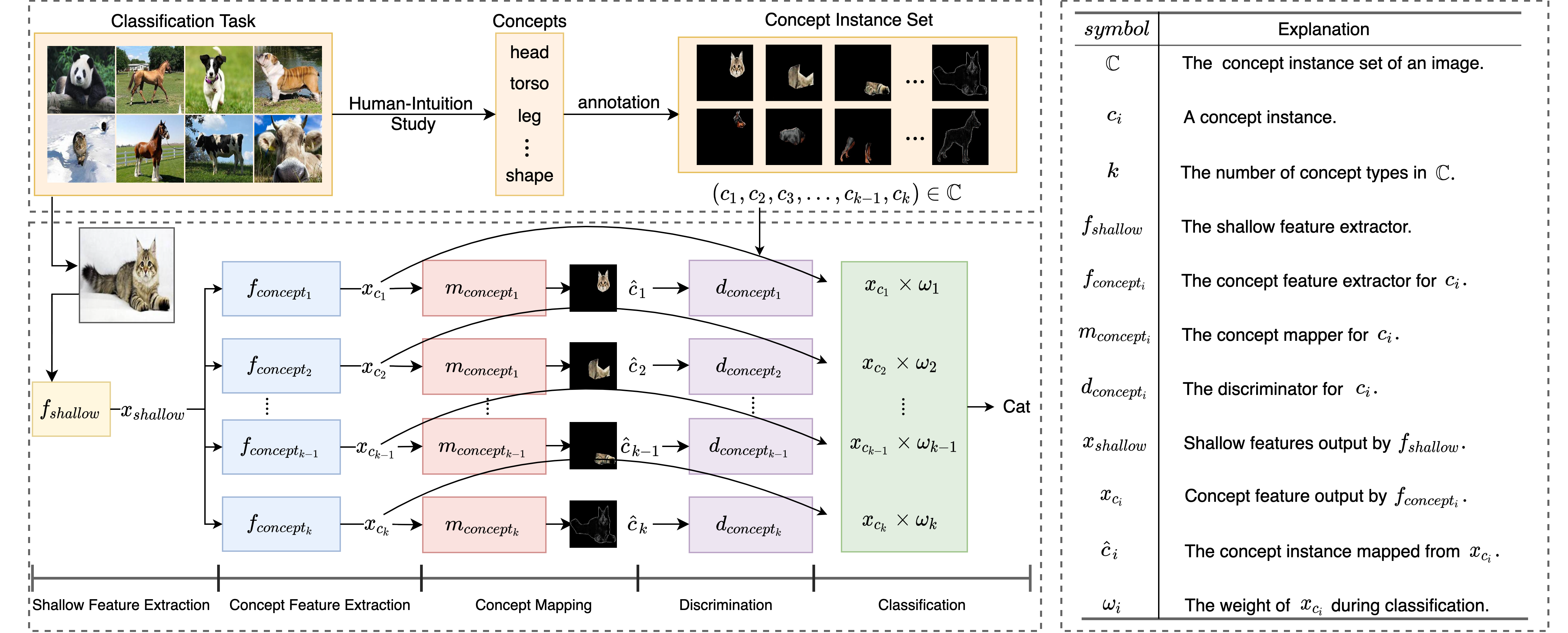}
    \caption{In TCNL, we first define some human-intuition concepts for specific classification tasks (this figure takes mammal classification task as an example). Then, images are fed into the shallow feature extractors to compute shallow features. From these shallow features, the concept extractor encodes specific concept-related information. The features output by all the concept extractors are concatenated and fed into the classifier for classification. At the same time, the concept mapper can map concept features to concept instances. The discriminator is used to classify concept instances mapped from concept features and original concept instances, which aims at improving the quality of concept learning.}
    \label{figure:overview}
\end{figure*}
 In order to ameliorate the issues mentioned above, we propose Transparent and Controllable Network Learning (TCNL), a novel approach to improve the transparency-interpretability and controllability of the CNN model. In TCNL, for specific tasks, we first define concepts corresponding to human understanding. Then we guide the model to learn disentangled knowledge from predefined concepts. Finally, the model accomplishes the classification task using features related to predefined concepts. The workflow of TCNL is presented in~\cref{figure:flow}.
 
 TCNL can be applied to existing CNN models such as VGG, ResNet, and AlexNet~\cite{vgg, alexnet, resnet}. TCNL improves interpretability and controllability by defining and learning concepts in accordance with human understanding for specific tasks. Meanwhile, TCNL is able to visualize concept information extracted by the model through the concept mapper. We specifically design an experiment to prove that the high-quality concept visualization stems from the successful concept learning process rather than a strong concept mapper. We will release all the datasets (including concept instance sets) in our experiments to support future scientific research on transparency-interpretability.

\section{Related Work}
\label{sec:related_work}

\textbf{Interpretable Models.}
Many studies focusing on transparency-interpretability have been carried out. Some works try to optimize the representation learning of neurons. Zhang \textit{et al.}~\cite{ICNN} try to train each filter in the high convolution layer to represent an object or a part. Based on the work in~\cite{ICNN}, Shen \textit{et al.}~\cite{icCNN} divide neurons into different groups in an unsupervised way to learn disentangled representations. However, connections between class labels and neurons are still entangled. To deal with this issue, Liang \textit{et al.}~\cite{CSGCNN} try to align each filter in the last convolution layer with a specific class during the learning process. Some approaches also try to improve the interpretability through structure adjustment. Nicola \textit{et al.}~\cite{part_whole} implement a neural network with a novel structure according to the visual cortex structure to represent the part-whole hierarchies and conceptual-semantic relationships. Pietro \textit{et al.}~\cite{entropy} propose an Entropy-based Network structure, trying to explain the model with First-Order Logic.

\textbf{Semantic Concepts.}
For semantic concepts, some methods pay attention to the concept found by the model during feature extraction. Zhou \textit{et al.}~\cite{detector} find that neurons in the deep layers attempt to detect a certain pattern or concept in the input image and they name these neurons as detectors. To quantitatively analyze the relationship between neurons and concepts, Zhou \textit{et al.}~\cite{netdissect} propose Network Dissection. Also, there are some methods focusing on finding important concepts for the prediction of the model. Kim \textit{et al.}~\cite{tcav} propose TCAV,  a novel framework to evaluate the importance of pre-set concepts to the decision of the model. However, TCAV requires additional training using pre-set concepts. To fill this gap, Amirata \textit{et al.}~\cite{ACE} propose the ACE algorithm to find important concepts for the decision of the model automatically leveraging the philosophy of unsupervised methods.

\textbf{Concept Bottleneck Models.}
The concept bottleneck model is a kind method that build connections between human understanding and model decision. The concept bottleneck model usually works in two steps. First, it extracts features from input data and predicts the concepts, then uses predicted concepts to make a final decision. The early version of the concept bottleneck model did not use neural networks, and it has been previously used for specific applications~\cite{cbm_face, cbm_animal}. More recently, the CNN technique has been merged into the concept bottleneck model for solving specific tasks~\cite{cbm_image_recog, cbm_retinal, cbm_vqa, cbm_classify}. Different from the concept bottleneck model, our work focuses on guiding the model to extract features related to predefined concepts rather than predicting the concepts.

\textbf{Visualization.}
Many methods have been proposed to visualize the decision of the CNN, knowledge learnt by the model, or the structure of the CNN. To explain the decision of the model, Zhou \textit{et al.}~\cite{cam} first propose the CAM algorithm to find and visualize the important regions of the input images that support the decision of the CNN. Along with the idea of the CAM algorithm, many CAM-based methods~\cite{gradcam, scorecam, relevancecam, cameras} have been proposed for better visualization and localization. For representation visualization, Dosovitskiy \textit{et al.}~\cite{inverting} propose the Inverting Network that can invert features to images. To visualize representations for neurons, Zhou \textit{et al.}~\cite{netdissect} propose a method based on image perturbation to visualize the Receptive Field and Activation Pattern of a single neuron.

\section{Method}
\label{proposed_method}
Our TCNL tries to make the process of feature extraction more understandable. In our TCNL, we first define some concepts following the logic of the human decision. Then we encourage the concept feature extractor to encode information related to predefined concepts. Based on the extracted concept features, the classifier makes decisions, and the concept mapper maps concept features to concept instances for visualizing representations of the concept feature extractor.

\subsection{Predefined Concepts and Datasets}
\begin{figure}[htbp]
    \centering
    \includegraphics[width=\linewidth]{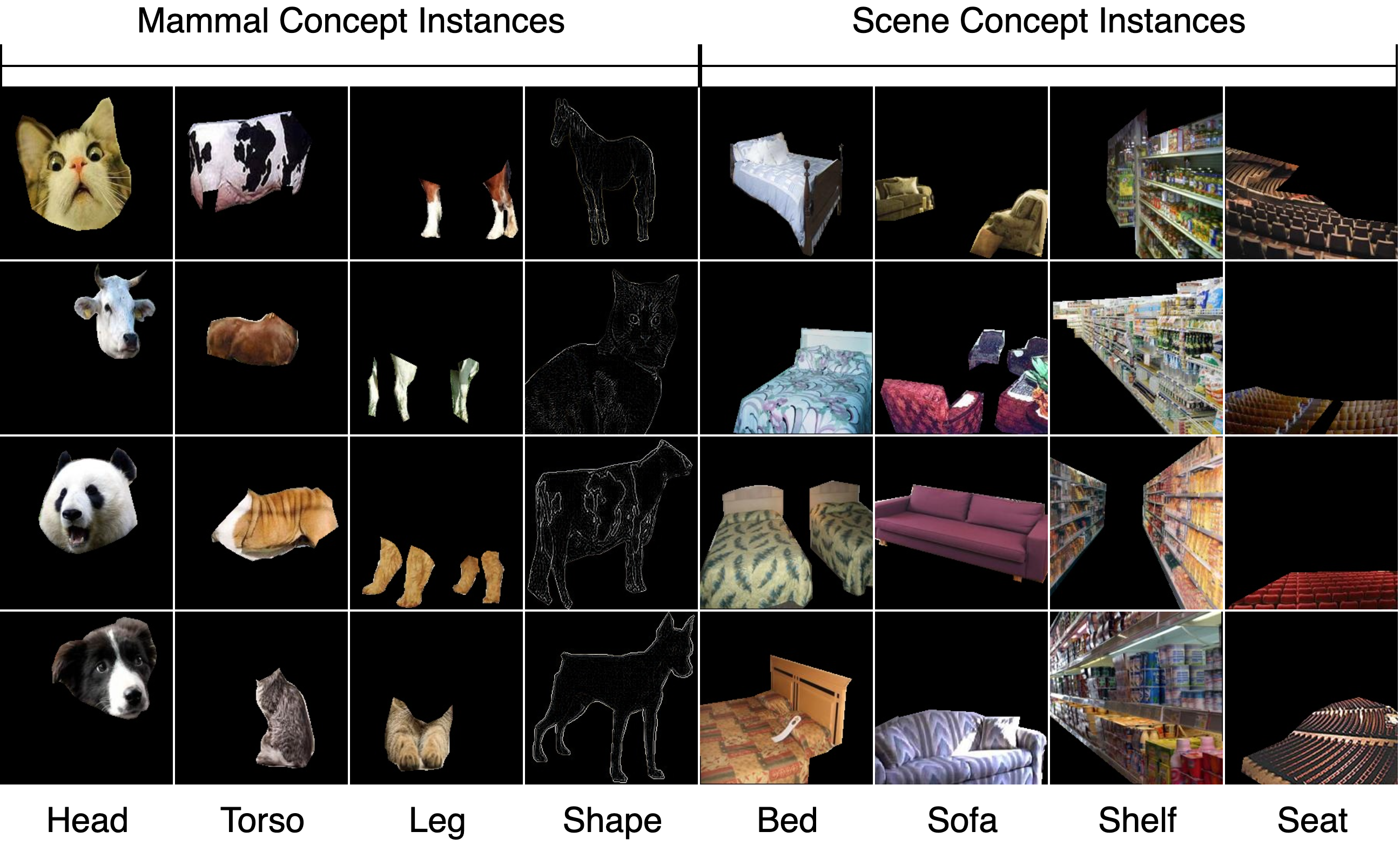}
    \caption{This figure shows some samples of the predefined concept instances. All the instances except the \textit{shape} instances are generated by pixel-wise dense annotation. The \textit{shape} instances are generated using the Laplacian operator.}
    \label{figure:concept_set}
\end{figure}

To define concepts in accordance with human understanding for specific tasks, we carry out a human-intuition study. 79 people participate in our study to define concepts for mammal classification task and scene classification task. According to the study result, we select different parts and the \textit{shape} of the mammal body as the key concepts for mammal classification. For scene classification, we find that different types of scenes may have totally different concepts. For example, the concept of the \textit{bed} may never appear in a theater scene. Therefore after defining concepts for scene classification, we also invite participators to select concepts that appeared in most images. Finally, we select \textit{head}, \textit{torso}, \textit{leg}, and \textit{shape} as concepts for mammal classification. For scene classification, we select \textit{bed}, \textit{sofa}, \textit{shelf}, and \textit{seat} as concepts. Examples of concept instances are shown in~\cref{figure:concept_set}. To reduce the bias, all the concepts are defined and selected by the 79 participators.
\begin{figure}
    \centering
    \includegraphics[width=\linewidth]{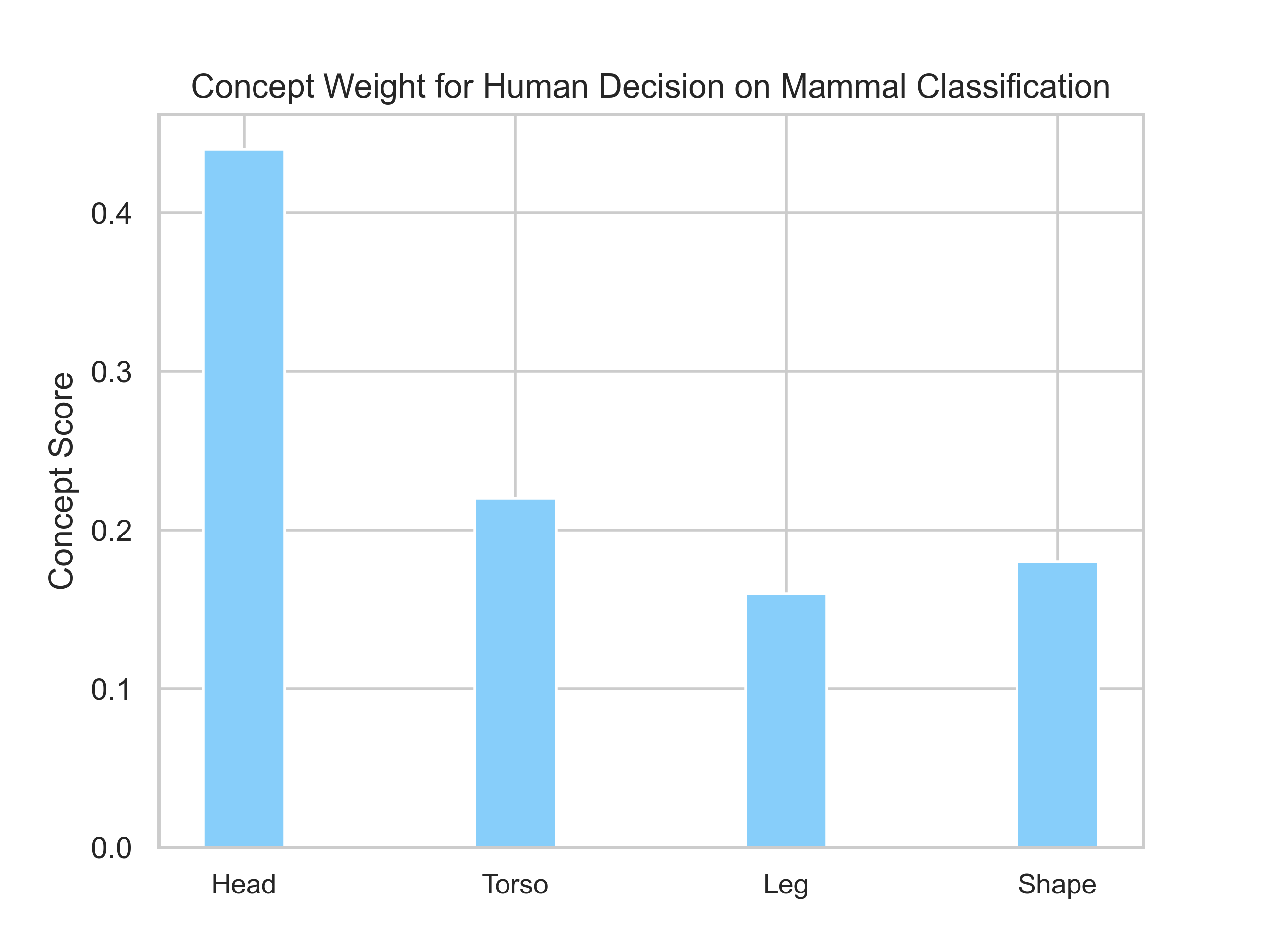}
    \caption{This figure shows the importance score of each concept for the human decision on the mammal classification task. The importance score is calculated based on the human-intuition study result.}
    \label{figure:mammal_concept_voting}
\end{figure}

As the concepts for mammal classification are shared between each class, we also invite people to sort the concepts we select according to the importance of these concepts in human decisions. The importance score for mammal concepts is shown in~\cref{figure:mammal_concept_voting}.

In TCNL, to guide the model to learn information about predefined concepts, we propose a mammal classification dataset and a scene classification dataset. With fine-grained annotation, we build concept instance sets for both datasets mentioned above. Considering most datasets~\cite{imagenet, coco, voc} do not match our method, we collect images from existing datasets~\cite{places} and the Internet to build our own datasets and give fine-grained annotation for every image to build concept sets. Mammal classification dataset includes 5 classes, which are cat, dog, cow, horse, and panda. Scene classification dataset includes 4 types of scenes in total, which are bedroom, living room, store, and theater.

\subsection{Structure and Learning}
The TCNL mainly aims at guiding the CNN to learn and encode information related to predefined concepts. An overview is shown in~\cref{figure:overview}. 

\textbf{Structure of TCNL.} In contrast to the traditional CNN structure, TCNL divides the CNN model into the shallow feature extractor, the concept feature extractor, the concept mapper, and the discriminator. Different parts of the model have different structures\footnote{The specific structure of each part can be adjusted according to the CNN model applied by TCNL} and perform different functions. The shallow feature extractor consists of some shallow convolutional layers and pooling layers, aiming at learning some simple features such as texture and color~\cite{netdissect}. The concept feature extractor contains some deep convolutional layers and it is encouraged to learn information about the predefined concepts from simple features. The concept mapper consists of some transposed convoluational layers. Based on concept features output by the concept feature extractor, the concept mapper maps features to images for visualizing concept representation learnt by the model. In addition, we use the discriminator to improve the performance of the model on concept learning.

\textbf{Feature Extraction.} TCNL guides the concept feature extractor to learn disentangled representations about predefined concepts. The feature extraction process in TCNL can be described using following formulas, and it is also shown in~\cref{figure:overview}.

First, in~\cref{formula:define concept}, we build the concept instance set for the specific task $T$. Concept instances are used as supervision information for concept learning.
\begin{equation}
    \begin{aligned}
    \resizebox{0.9\linewidth}{!}{$T\xrightarrow{}\{(c_1^1, \dots, c_1^{l-1}, c_1^l), \dots, (c_k^1, \dots, c_k^{l-1}, c_k^l)\} \in \mathbb{C} \text{,}$}
    \end{aligned}
    \label{formula:define concept}
\end{equation}
\noindent where $c_i^j$ denotes a certain concept instance from concept $c_i$ and $\mathbb{C}$ denotes a concept instance set including $k \times l$ instances ($k$ denotes the number of the concepts and $l$ denotes the number of instances from a certain concept). For the clear expression of the formulas, we use $c_i$ to denote an instance from a specific concept in the following content.

Second, in~\cref{formula:feature extraction}, the shallow feature extractor $f_{shallow}$ computes the shallow feature $x_{shallow}$ of the input image $I$. Finally, $x_{shallow}$ is passed to the concept feature extractor to compute the concept feature $x_{c_i}$ related to concept $c_i$. 
\begin{equation}
    \begin{aligned}
        x_{c_i} = f_{c_i}(x_{shallow}) = f_{c_i}(f_{shallow}(I)) \text{,}
    \end{aligned}
    \label{formula:feature extraction}
\end{equation}
\noindent where $f_{shallow}$ and $f_{c_i}$ denote the shallow feature extractor and the concept feature extractor, respectively. $x_{shallow}$ and $x_{c_i}$ represent the output features of the shallow feature extractor and the concept feature extractor, respectively.

After the feature extraction, the concept mapper maps the concept feature to the concept instance $\hat{c}_i$ for visualization and the classifier makes the final decision.

\textbf{Concept Learning.} TCNL encourages the model to encode concept-related information while keeping the outstanding performance on classification using the constraint in~\cref{formula:whole loss}. $Loss_{gan}$ and $Loss_{similarity}$ aim at concept learning. $Loss_{classification}(\hat{y},y)$ is a cross entropy loss to keep the classification performance.
\begin{small}
\begin{equation}
    \begin{aligned}
    \resizebox{0.9\linewidth}{!}{$Loss = \lambda Loss_{gan} + \mu Loss_{similarity} + \eta Loss_{classification}(\hat{y},y) \text{.}$}
    \end{aligned}
    \label{formula:whole loss}
\end{equation}
\end{small}

Towards the goal of guiding the model to learn knowledge from predefined concepts, we use $Loss_{similarity}$ and $Loss_{gan}$ in~\cref{formula:whole loss} to constrain the learning process. For $Loss_{similarity}$, it can be described as~\cref{formula:loss_similarity}. $Loss_{similarity}$ is calculated at the end of the concept mapper, and affects the concept feature extractor and the concept mapper. It measures the pixel-wise mean square error between the original concept instance $c_i$ and the visualized concept instance $\hat{c}_i$. With this constraint, the concept feature extractor is guided to encode concept-related information and the concept mapper is encouraged to map concept features to concept instances.
\begin{equation}
\begin{aligned}
    Loss_{similarity} = \frac{1}{w \times h }\sum_{n}^{w} \sum_{m}^{h}(c_{nm}-\hat{c}_{nm} )^2 \text{,}
\end{aligned}
    \label{formula:loss_similarity}
\end{equation}
\noindent where $w$ and $h$ denote the width and height of the concept instance, respectively. $c_{nm}$ and $\hat{c}_{nm}$ represent the pixel at coordinates $(n,m)$ in concept instance $c_i$ and visualized concept $\hat{c_i}$.

We leverage the advantages of GAN in our TCNL to further enhance the ability of the model to learn predefined concepts. $Loss_{gan}$ in~\cref{formula:whole loss} can be described as~\cref{formula:loss_gan}. Consistent with the philosophy of GAN, a discriminator is used to classify the original concept instance $c_i$ and visualized concept instance $\hat{c}_i$. Under the influence of the discriminator, the concept feature extractor and the concept mapper can have better performance in encoding and mapping concept features.
\begin{equation}
\begin{aligned}
\resizebox{0.9\linewidth}{!}{$ Loss_{gan}= \mathbb{E}_{\hat{c}_i}[logD(\hat{c}_i)]+ \mathbb{E}_{c_i, \hat{c}_i}[log(1-D(c_i, \hat{c}_i))] \text{,}$}
\end{aligned}
\label{formula:loss_gan}
\end{equation}
\noindent where $D$ denotes the discriminator. Discriminator $D$ tries to maximize this function while other parts of the model with TCNL try to minimize it.

\begin{figure*}[htb]
    \centering
    \includegraphics[width=0.9\linewidth]{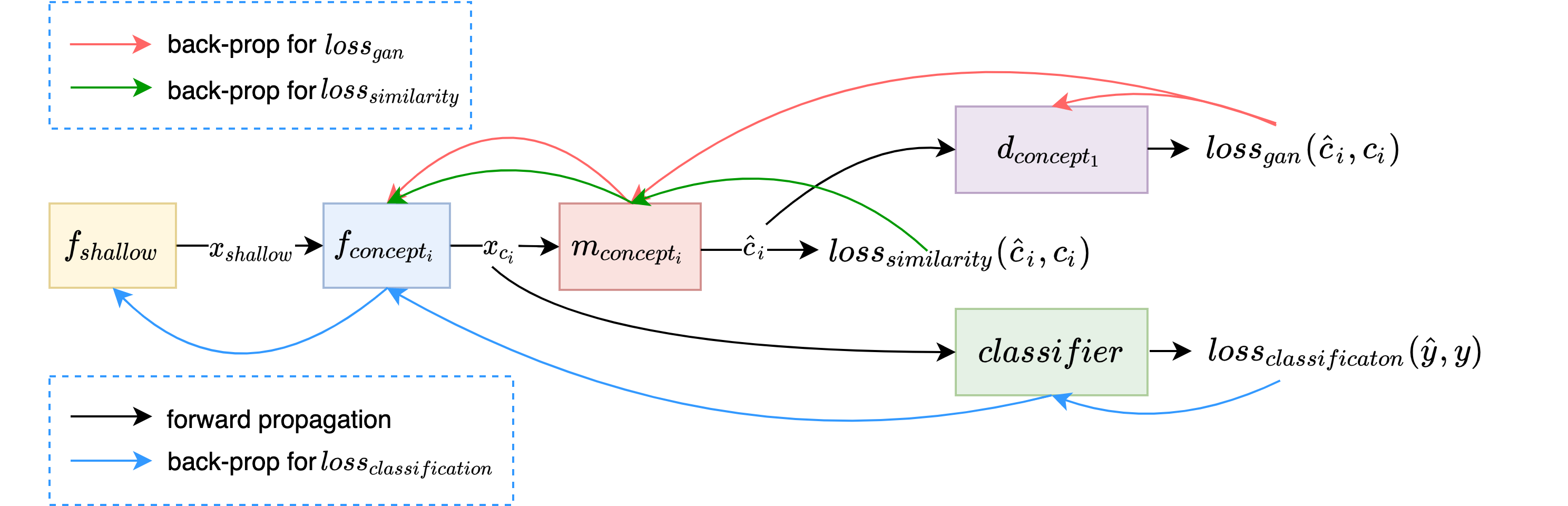}
    \caption{This figure shows the backward propagation process in our TCNL during training.}
    \label{figure:backward}
\end{figure*}

The backward process is presented in~\cref{figure:backward}. $Loss_{similarity}$ is back-propagated to the concept feature extractor and the concept mapper. $Loss_{gan}$ is back-propagated to the concept feature extractor, the concept mapper, and the discriminator. $Loss_{classification}(\hat{y},y)$ is back-propagated the shallow feature extractor, the concept feature extractor, and the classifier. During the backward propagation, the discriminator and other parts of the model are optimized separately.

\section{Experiments}
\label{experiments}

\subsection{Datasets and Implementation}
\textbf{Datasets:} All the experiments are performed on mammal classification dataset and scene classification dataset. Mammal classification dataset contains 5 classes, 2500 mammal images (500 images for each class), and 10000 concept instances (2000 instances for each class). Scene classification dataset contains 4 classes, 2000 scene images (500 images for each class), and 8000 concept instances (2000 instances for each class). Some samples of the concept instances are shown in~\cref{figure:concept_set}.

\textbf{Implementation.} As has been done in other studies~\cite{entropy, ICNN, netdissect, inverting}, we apply our TCNL to three traditional CNN models (VGG, AlexNet, ResNet). For hyper-parameters of the training process, we set the learning rate to 0.001, and the batch size to 8 for both datasets. 

\subsection{Metrics}
For evaluating interpretability, we propose Concept-Related Neuron Proportion (CRNP\footnote{The definition of CRNP is shown in~\cref{crnp}}), which represents the proportion of neurons that are sensitive to a certain concept. Higher CRNP means more neurons tend to encode information from a certain concept. We use Mean Squared Error (MSE) and Structural Similarity (SSIM)~\cite{ssim} as the evaluation indicators for the performance of the concept mapper. MSE measures the pixel-wise similarity between original concept instances and visualized concept instances, and SSIM comprehensively measures the differences in image brightness, contrast, and structure. For MSE metric, lower is better. For SSIM metric, higher is better. We also use Accuracy (ACC) to measure the performance on classification tasks. 

\subsection{Results}
\begin{figure*}[htb]
    \centering
    \includegraphics[width=0.9\linewidth]{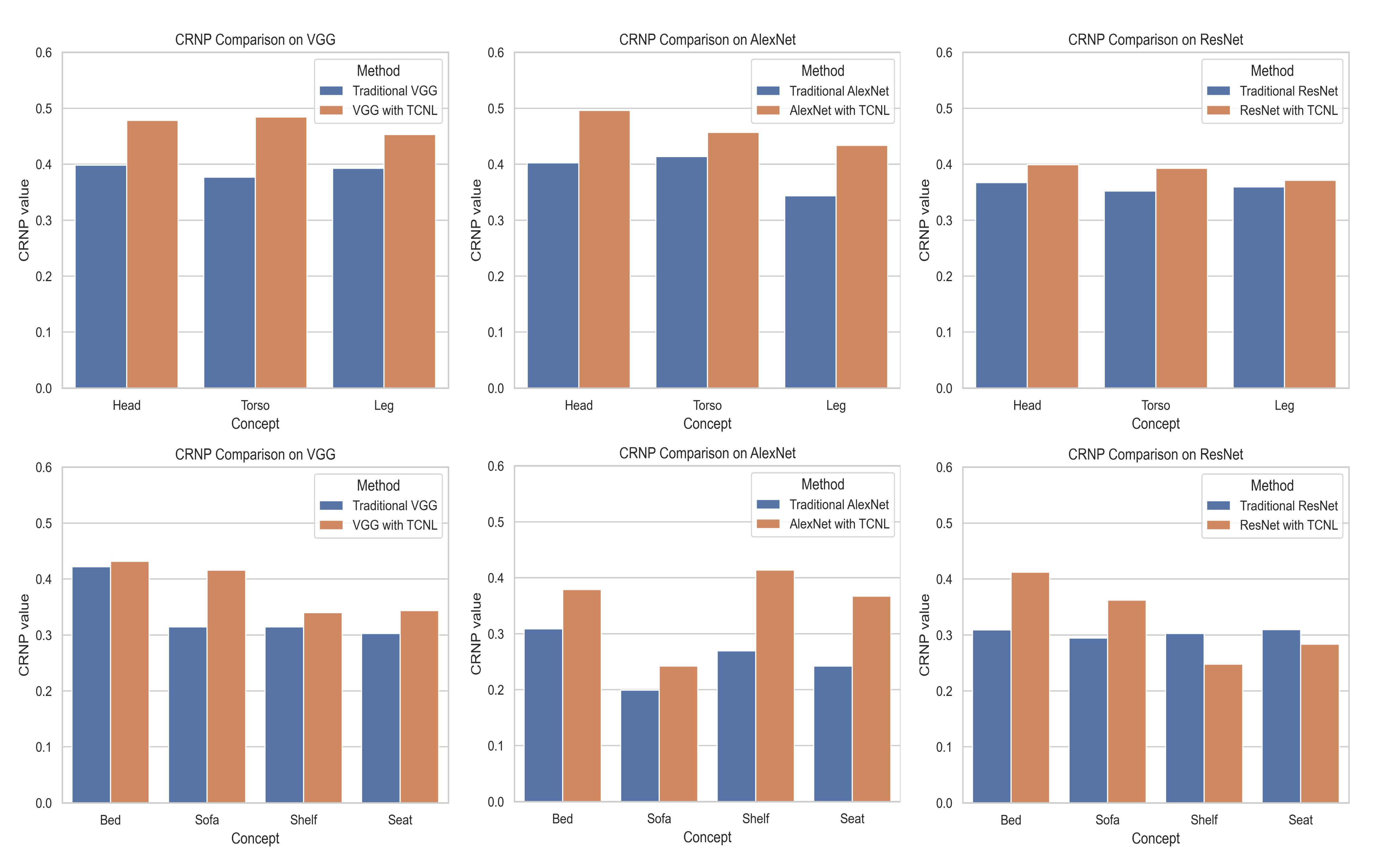}
    \caption{This figure shows the CRNP result on each concept in mammal classification dataset and scene classification dataset. We compare the result between traditional CNNs (VGG, ResNet, and AlexNet) and models with TCNL.}
    \label{figure:crnp_comparison}
\end{figure*}
\label{result}
\subsubsection{Concept-Related Neurons Analysis}
\label{crnp}
\ 
\indent Existing research~\cite{detector, netdissect} shows that neurons in deep layers tend to detect high-level concepts like objects and parts in the image. When concepts are removed from the image, the activation value usually drops. Zhou \textit{et al.}~\cite{detector} use the numerical drop of the activation of neurons to define \textbf{Detectors}. Similarly, we define Concept-Related Neuron . Taking the \textit{head} concept as an example, we first calculate the activation value of each neuron in the last layer of the head concept feature extractor using the full image as input. Then we remove the head part from the image and calculate the activation value again using the new image as input. Finally, calculate the average numerical drop of all the neurons in the last layer of the head concept extractor. Neurons whose activation value decreases more than the average numerical drop are defined as Concept-Related Neurons. The proportion of the Concept-Related Neurons is named as CRNP.

To analyze the transparency-interpretability of our TCNL, we calculate the proportion of concept-related neurons in the last layer of the concept feature extractor on mammal classification dataset. As the result in~\cref{figure:crnp_comparison} shows, our TCNL has a better performance on CRNP. On every concept in the mammal classification task, all the models with TCNL outperform traditional CNNs. For the scene classification task, VGG with TCNL and AlexNet with TCNL both outperform traditional CNNs on every concept. However, for ResNet with TCNL, it does not perform as well as the traditional ResNet on the \textit{shelf} concept and the \textit{seat} concept. This may be because, in addition to the concept constraints, we also use the classification constraint to ensure the discriminating ability. The classification constraint may affect the ability of the model to learn specific concepts.

\subsubsection{Concept Weight Analysis}
\ 
\indent To further analyze the transparency-interpretability of TCNL, we quantitatively measure the importance of the concepts during the decision process of the model on mammal classfication dataset. To calculate weight for these concepts, we use gradient back-propagated to the corresponding concept feature as the concept weight, which is in accordance with the weight calculating method in~\cite{gradcam}. The gradient value measures the sensitivity of a concept feature to the decision of the model. A higher gradient value represents a higher weight in classification.
\begin{figure*}[htbp]
    \centering
    \includegraphics[width=0.9\linewidth]{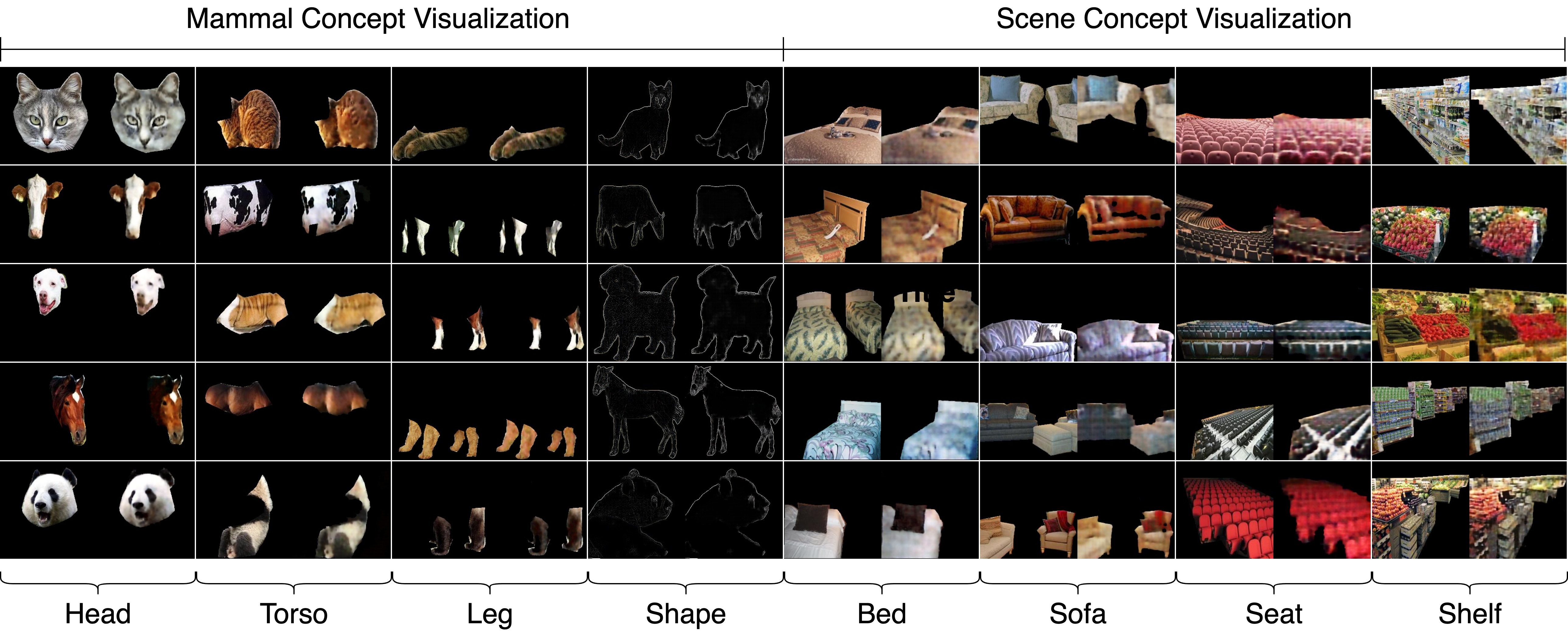}
    \caption{This figure shows the comparison between instances visualized from the learnt concept features and the original concept instances. For each image pair, the left one is the original concept instance accessed through annotation and the right one is the output of a concept mapper.}
    \label{figure:reconstruction_vis}
\end{figure*}
According to the result in~\cref{table:mammal_concept_weight}, for human decision, the importance rank of concepts is \textit{head}, \textit{torso}, \textit{shape}, and \textit{leg}. However, in our TCNL, four types of concepts (\textit{head}, \textit{torso}, \textit{leg}, and \textit{shape}) have similar weights and the \textit{shape} concept gets the biggest weight in the decision of the model. The reason for the difference is that we only impose concept constraints on the feature extraction stage. For the classifier, we only require it to have a good classification performance, which may cause the difference in concept weights.
\begin{table}[htb]
\begin{center}
\resizebox{\linewidth}{!}{
\begin{tabular}{ccccccccccccccccccccccc}
\hline
Subject of decision-making  &Head  &Torso  &Leg  &Shape \\ \hline
VGG with TCNL    &0.22  &0.22  &0.23  &0.33  \\
Human  &0.44  &0.22  &0.16  &0.18\\
\hline
\end{tabular}
}
\end{center}
\caption{Concept weights on mammal classification task.} 
\label{table:mammal_concept_weight}
\end{table}

\begin{table*}[htb]
\begin{center}
\resizebox{0.9\textwidth}{!}{
\begin{tabular}{c|c|cccccccccccccccc}
\hline
\multicolumn{2}{c|}{Model and Metric}  &Head  &Torso  &Leg  &Shape  &Bed  &Sofa  &Shelf  &Seat\\ \hline
\multirow{2}{*}{ResNet-50 with TCNL} &MSE    &44.44  &36.52  &31.34  &276.49  &93.38  &50.61  &136.75  &64.44\\
&SSIM   &0.96  &0.96  &0.98  &0.74  &0.95  &0.94  &0.92  &0.95\\ \hline
\multirow{2}{*}{VGG-11 with TCNL} &MSE    &54.82  &48.03  &26.47  &205.06  &44.79  &35.97  &105.58  &69.28\\
&SSIM   &0.96  &0.95  &0.98  &0.77  &0.96  &0.97  &0.94  &0.94\\ \hline
\multirow{2}{*}{AlexNet with TCNL} &MSE    &98.98  &111.56  &115.04  &505.09  &126.26  &192.80  &152.45  &230.87\\
&SSIM   &0.92  &0.92  &0.93  &0.69  &0.96  &0.93  &0.92  &0.92\\ \hline
\end{tabular}
}
\end{center}
\caption{Evaluation of the visualization quality on each concept using MSE and SSIM.} 
\label{table:result_reconstruction}
\end{table*}

\subsubsection{Visualization Analysis} 
\ 
\indent To evaluate the performance of the visualization, we first train three types of models (VGG, AlexNet, ResNet) with TCNL on mammal classification dataset and scene classification dataset. Then, we collect the concept visualization results of each image on the two datasets. For each concept, we calculated MSE and SSIM to evaluate the performance of the concept mapper. The visualization results presented in~\cref{table:result_reconstruction} and~\cref{figure:reconstruction_vis} prove that the concept mapper can successfully map concept features to concept instances based on the concept representation of the model.
\begin{figure}[htb]
    \centering
    \includegraphics[width=\linewidth]{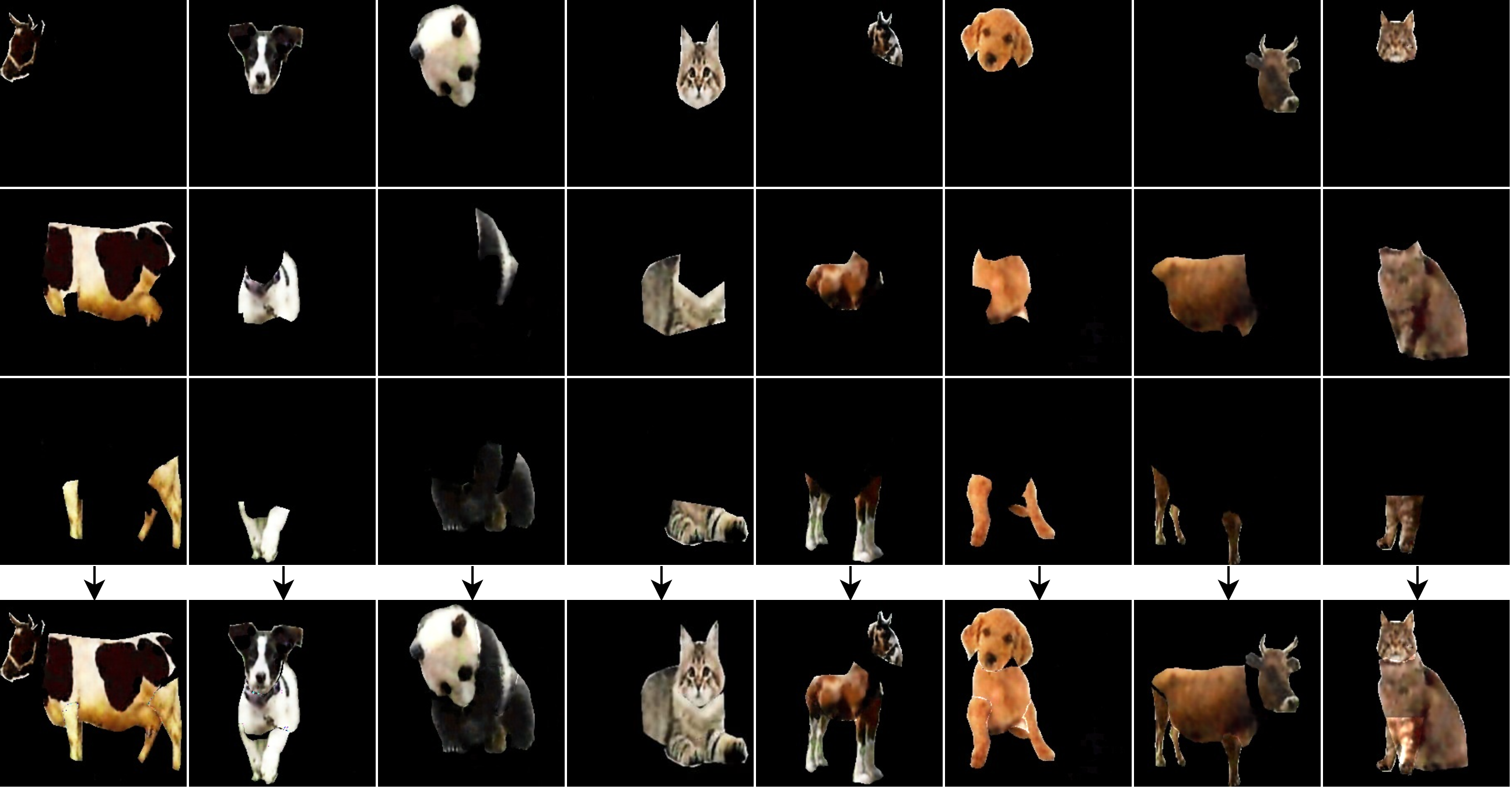}
    \caption{This figure shows the positional association of the concept instances. Images above the arrow are concept instances visualized by the concept mapper. Images below the arrow are concatenated from individual concept instances.}
    \label{figure:postion_vis}
\end{figure}

\begin{table}[htb]
\begin{center}{
\resizebox{\linewidth}{!}{
\begin{tabular}{c|c|ccccccccc}
\hline
\multicolumn{2}{c|}{Method and Metric}  &Head  &Torso  &Leg  &Shape\\ \hline
\multirow{2}{*}{VGG-11 without Concept Constraint} &MSE    &222.67  &113.50  &95.07  &357.95  \\
&SSIM   &0.90  &0.91  &0.94  &0.74\\ \hline
\multirow{2}{*}{VGG-11 with Concept Constraint} &MSE    &54.82  &48.03  &26.47  &205.06\\
&SSIM   &0.96  &0.95  &0.98  &0.77 \\
\hline
\end{tabular}}
}
\end{center}
\caption{Visualization comparison between the complete TCNL method and TCNL without concept-related constraint.} 
\label{table:comparision_similarity}
\end{table}

We also concatenate concept instances visualized by the concept mapper to analyze the positional association of the concept instances. The result in~\cref{figure:postion_vis} shows that TCNL is also able to help the model learn position information among concept instances.
\begin{figure*}[htb]
    \centering
    \includegraphics[width=0.85\linewidth]{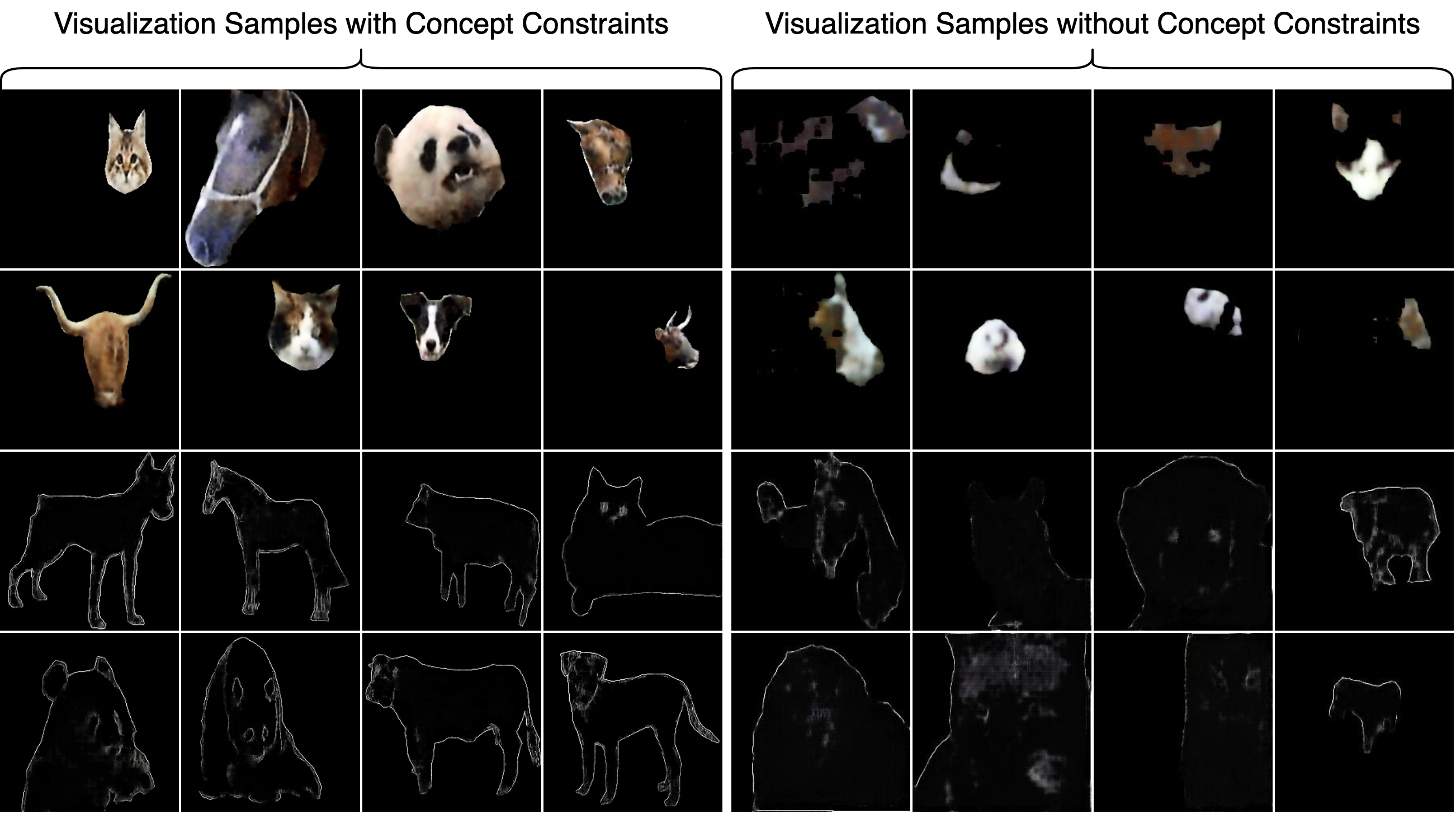}
    \caption{In this figure, we compare the visualization result between the complete TCNL method and TCNL without the concept-related constraint for concept feature extractors.}
    \label{figure:feasibility}
\end{figure*}

\subsubsection{Validating the Concept Learning}
\ 
\indent To demonstrate that the high-quality concept visualization stems from concept knowledge learned by the model, rather than a powerful concept mapper, we specifically design this contrast experiment. We applied our TCNL on two same VGG models. The first model does not have the concept-related constraint for encoding concept information while other parts of the model are the same as we have proposed in~\cref{proposed_method}. The second model is trained with complete TCNL method. These two models are trained on our mammal classification dataset with the same training hyper-parameters (Batch size set to 8, learning rate set to 0.001). Then we evaluate the visualization performance of the concept mapper using MSE and SSIM.

As the result in~\cref{table:comparision_similarity} shows, the model with the concept-related constraint gives a better performance. We also present visualization result of these two models in~\cref{figure:feasibility}. It is clear that the concept constraint in TCNL helps the model better learn knowledge about predefined concepts.
It is clear that the model with our TCNL successfully learns and visualizes the predefined concepts.

\subsubsection{Classification Performance}
\label{result_normal_exp}
\begin{table}[htb]
\begin{center}
\resizebox{\linewidth}{!}{
\begin{tabular}{c|c|cccccccccccccccccccccc}
\hline
\multicolumn{2}{c|}{Dataset and Method}  &VGG-11  &ResNet-50  &AlexNet \\ \hline
\multirow{2}{*}{Mammal} &Original Method    &0.68  &0.70  &0.76\\
&TCNL Method  &\textbf{0.82}  &\textbf{0.74}  &\textbf{0.82} \\ \hline
\multirow{2}{*}{Scene} &Original Method    &\textbf{0.82}  &\textbf{0.80}  &0.78 \\ 
&TCNL Method   &0.80  &0.75  &\textbf{0.83}\\ \hline
\end{tabular}}
\end{center}
\caption{Classification performance comparison between models with TCNL and traditional CNNs on mammal classification dataset and scene classification dataset using ACC.} 
\label{comparision_acc}
\end{table}

In addition, we also evaluate the classification performance of the model with TCNL. We train three types of CNN models (VGG, ResNet, AlexNet) with TCNL on mammal classification dataset and scene classification dataset. At the same time, we also train traditional VGG, ResNet, and AlexNet, which are used as baseline methods. ACC is used to measure classification performance. The result in~\cref{comparision_acc} shows that our TCNL does not have a serious adverse effect on the classification performance. Models with TCNL even perform better than traditional CNN models on some tasks. It is clear that, in addition to the better interpretability, TCNL maintains a strong discriminating ability as the traditional CNN and achieves a balance between interpretability and discriminating power. 

\section{Conclusion}
In this paper, we propose TCNL to guide the model to learn knowledge about the predefined concepts. Therefore, transparency-interpretability of the model is improved. In our method, concepts (such as \textit{head}, \textit{leg}, \textit{bed}, \textit{sofa} and so on) that fit the logic of the human decision can be defined artificially. In TCNL, the model is divided into the shallow feature extractor, the concept feature extractor, the concept mapper, the discriminator, and the classifier. Concept instances used for concept learning can be easily accessed through artificial annotation. With the concept-related constraint in TCNL, the concept feature extractor is guided to encode information related to predefined concepts and the concept mapper is encouraged to map concept features to concept instance images. Referring to the successful utility of our method, we expect that TCNL has the potential to help people understand and gain more control on the CNN in more areas than the classification task. 

We should notice that our TCNL relies on annotated data to some extent. To support related research, we annotated animal data and indoor scene data. All the data (including concept instance sets), code and following work will be released for scientific research, and all the suggestions and contribution are welcomed.

{\small
\bibliographystyle{ieee_fullname}
\bibliography{egbib}
}

\end{document}